\documentclass[11pt]{article}

\usepackage[preprint]{acl}

\usepackage{times}
\usepackage{latexsym}

\usepackage[T1]{fontenc}
\usepackage[utf8]{inputenc}

\usepackage{microtype}

\usepackage{inconsolata}

\usepackage{graphicx}
\usepackage{booktabs}
\title{Accounting Reasoning in Large Language Models: Concepts, Evaluation, and Empirical Analysis}

\author{Jie Zhou,  Xin Chen, Jie Zhang, Zhe Li \\
School of Computer Engineering, Jiangsu Ocean University \\
jzhou23@jou.edu.cn, lizhe@jou.edu.cn
}

\begin{document}
\maketitle
\begin{abstract}
Large language models (LLMs) are increasingly reshaping learning paradigms, cognitive processes, and research methodologies across diverse domains. As their adoption expands, effectively integrating LLMs into professional fields and clarifying their role in domain-specific applications has become a key challenge for enterprise digital transformation and broader societal development. In the accounting domain, successful integration requires a systematic understanding of LLMs’ domain-specific reasoning capabilities.
In this study, we introduce the concept of accounting reasoning and propose a set of evaluation criteria grounded in an analysis of the training data characteristics of representative GLM-series models. These criteria establish a foundation for studying accounting-oriented reasoning paradigms and provide benchmarks for assessing and improving model performance. Building on this framework, we evaluate several representative LLMs, including GLM-6B, GLM-130B, GLM-4, and GPT-4, across a range of accounting reasoning tasks.
Our experimental results show that prompt engineering strategies can yield varying degrees of performance improvement across models, with GPT-4 demonstrating the strongest overall accounting reasoning capability. Nevertheless, the results indicate that current LLMs remain insufficient for real-world accounting applications. In particular, further optimization is required for deployment in enterprise-level accounting scenarios to fully realize the potential value of LLMs in this domain.
\end{abstract}

\section{Introduction}

With the rapid development of artificial intelligence technologies, fields such as big data, supercomputing, brain-inspired intelligence, and large language models (Large Language Models, LLMs), particularly general-purpose natural language models, have attracted widespread attention from both academia and industry. Artificial intelligence has become a new engine for social intelligence and a core driving force for a new round of technological innovation and industrial transformation. It is also a key driver of national innovation capacity and an important strategic component of national security. In 2017, the State Council issued the New Generation Artificial Intelligence Development Plan, which identified artificial intelligence as a major strategic priority, emphasized the importance of strengthening China’s advantages in the AI field, and promoted interdisciplinary exploratory research. The plan encouraged the integration of artificial intelligence with mathematics, engineering, and other disciplines, strengthened foundational research, and supported the application of AI technologies across various sectors.

In July 2023, multiple national ministries jointly issued the Interim Measures for the Administration of Generative Artificial Intelligence Services, encouraging innovation and application of generative artificial intelligence technologies across industries and domains, and promoting the establishment of a standardized, orderly, and innovative governance framework. This policy context has created favorable conditions for the integration of LLMs with accounting practice.

As LLMs continue to evolve, they are no longer limited to foundational computing power or model-scale competition, but are increasingly driving the emergence of intelligent industries characterized by deep application integration. With the rapid deployment of general-purpose LLMs in vertical domains, how to effectively integrate LLMs into specific professional workflows and release their domain-specific potential has become a key direction for future research and technological development. Compared with narrowly specialized models, general-purpose LLMs demonstrate strong advantages in accounting-related tasks, including richer cross-domain knowledge, enhanced prompt understanding capabilities, and flexible reasoning and inference processes. These strengths enable LLMs to be applied across a wide range of accounting scenarios, such as intelligent financial analysis, tax compliance, auditing support, and management decision making. However, despite rapid exploration and experimentation, effective domain-specific LLMs for accounting have not yet been systematically established, and the practical effectiveness of LLM-based accounting systems remains limited.

In existing applications that integrate LLMs with accounting tasks, prompt engineering is commonly adopted as the primary interaction paradigm. This approach enhances model performance by constructing structured prompts or task-specific templates. However, such methods lack in-depth analysis of the internal reasoning mechanisms of LLMs and are therefore difficult to specialize or optimize systematically. Both approaches rely primarily on the general capabilities of LLMs and do not fundamentally address the need for domain-adaptive reasoning. From the perspectives of mathematical reasoning, logical inference, and symbolic computation, the accounting domain places particularly high demands on the reasoning capabilities of LLMs.

Based on insights from natural language generation (NLG) and natural language understanding (NLU), this study argues that the core computational capability underlying LLM performance in accounting tasks is reasoning ability. Accordingly, we propose to systematically analyze and evaluate the accounting reasoning capabilities of LLMs. By combining accounting professional knowledge with LLM reasoning paradigms, we construct an evaluation dataset tailored to accounting expertise and analyze experimental results to identify directions for further optimization.

The main contributions of this paper are as follows. First, we extend research on general-purpose LLMs to accounting reasoning. We analyze the relationships between LLM reasoning ability, mathematical reasoning, logical inference, and accounting expertise, providing a theoretical foundation for subsequent studies. Second, we design a comprehensive and diverse benchmark dataset for accounting reasoning evaluation. Unlike existing datasets that focus on single financial subdomains, our dataset covers accounting principles, financial statements, cost accounting, and auditing, enabling a more comprehensive assessment of LLM accounting reasoning performance. Third, we evaluate representative LLMs, including GPT-4 and GLM-series models, on the proposed benchmark and analyze their strengths and limitations, offering valuable insights for accounting-related applications. Finally, we explore the integration of domain knowledge with reasoning optimization strategies to improve LLM performance in accounting tasks, providing guidance for future research and accelerating the deployment of intelligent accounting systems in real-world practice.

\section{Related Work}

\subsection{Development of LLMs in the Accounting Domain}

The successful application of general-purpose large language models (LLMs) in vertical domains such as medicine and law provides important references for their deployment in the accounting domain. In medical applications, LLMs have been used in clinical decision support, medical image interpretation, and medical record generation \citep{panagiotou2024}. Due to the high interpretability of medical imaging data, LLMs can effectively identify visual patterns and, when combined with natural language generation, improve the transparency and explainability of AI systems, leading to better clinical outcomes \citep{chalkidis2023}.

In the legal domain, LLMs are often deployed at the application level, as legal environments constitute complex human-designed systems characterized by uncertainty and strict regulatory constraints. AI assistance in this domain must comply with legal norms and ethical requirements, which significantly limits the scope of automated reasoning. The accounting domain shares similarities with law in that both rely on explicit rules and structured standards. However, accounting places greater emphasis on quantitative reasoning, numerical accuracy, and consistency across financial records. Unlike medical image-based tasks, accounting tasks require LLMs to retain accounting standards, reason over causal relationships, and perform multi-step logical and numerical inference. As a result, reasoning capability is particularly critical for accounting applications, especially multi-step computational reasoning.

Existing research mainly focuses on the following directions. First, some studies aim to improve specific reasoning abilities of LLMs, such as mathematical or logical reasoning \citep{huang2022reasoning}. However, these methods often struggle to integrate accounting-specific knowledge, as they assume that reasoning capability can be enhanced independently of domain expertise. Consequently, such approaches show limited effectiveness in accounting scenarios \citep{li2023}. Moreover, there remains a lack of work explicitly targeting accounting reasoning.

Second, some studies attempt to improve domain performance by specializing training data or scaling model parameters. Models such as BloombergGPT \citep{wu2023bloomberggpt} and FinGPT \citep{liu2023fingpt} incorporate large-scale financial corpora or increase model capacity to enhance performance. However, these approaches do not necessarily lead to stable improvements in reasoning ability and often lack interpretability, making it difficult to assess whether performance gains stem from genuine reasoning improvements or surface-level pattern matching \citep{zhao2023}.

Third, evaluation in some works primarily focuses on answer accuracy. For example, LLMs are evaluated on tasks such as financial question answering or accounting knowledge retrieval \citep{faturos2023,theuma2024,shah2023}. These tasks are largely text-based and do not reflect the structured, multi-step nature of accounting reasoning. As a result, they fail to capture the core reasoning capabilities required in accounting applications.

Fourth, several exploratory studies discuss the potential of LLMs in accounting practice \citep{he2023,ouyang2024}. While these works provide valuable insights, they have not yet established systematic research paradigms or concrete optimization strategies for improving accounting reasoning capabilities.

Overall, existing studies on LLM applications in accounting primarily focus on general language understanding tasks such as classification, reading comprehension, and sentiment analysis. There remains a lack of systematic benchmarks and evaluation frameworks that explicitly target accounting reasoning.

\subsection{Reasoning Capability of LLMs in Accounting}

In LLM research, reasoning is commonly defined as the ability to infer conclusions from given facts and constraints \citep{huang2022reasoning,yu2023}. Logical reasoning evaluations typically assess the correctness of model outputs on inference tasks. According to reasoning structure, reasoning can be categorized into mathematical reasoning, logical reasoning, causal reasoning, and commonsense reasoning \citep{sun2023,huang2022reasoning}. Among these categories, mathematical reasoning and logical reasoning are most closely related to accounting tasks.

Accounting problems often involve complex numerical computation, multi-step logical deduction, and strict rule-based constraints. Therefore, reasoning capability plays a central role in determining LLM performance in accounting applications. Logical reasoning evaluates whether models can derive valid conclusions based on a set of premises and rules, typically measured through natural language inference tasks \citep{srivastava2022,yu2023,chang2023}. However, existing inference benchmarks are designed for general language understanding and do not adequately reflect the structured reasoning demands of accounting.

Mathematical reasoning evaluates a model's ability to perform arithmetic and symbolic computation. This capability is commonly assessed using datasets such as GSM8K \citep{cobbe2021}, which consist of multi-step elementary math word problems. These tasks require sequential reasoning and intermediate result propagation, which closely resemble accounting calculation processes.

In accounting contexts, except for high-level financial analysis, most tasks involve multi-step numerical computation. If LLMs fail to correctly propagate intermediate results across reasoning steps, errors accumulate and lead to incorrect conclusions. Therefore, multi-step mathematical reasoning is particularly critical. 
Moreover, many accounting scenarios require reasoning over temporally evolving entities and constraints, where integrating temporal knowledge graphs \citep{xiongtilp, xiong2024teilp, xiong2024large} with LLM-based symbolic reasoning frameworks \citep{yang2024harnessing} can provide structured state tracking and improve consistency across multi-step inference.
SWAP \citep{xiong2025deliberate} introduces structure-aware planning to enhance multi-step reasoning performance. 
MR-GSM8K \citep{zeng2024} extends GSM8K by filtering problems with short reasoning chains and increasing reasoning depth, resulting in longer and more complex reasoning trajectories, resulting in longer and more complex reasoning trajectories that better resemble real-world accounting calculations.
After further filtering, 586 MR-GSM8K problems requiring at least three computation steps remain. This subset can serve as a foundational benchmark for evaluating multi-step calculation ability in accounting contexts, referred to as the Multi-Calculation Benchmark.

\subsection{Evaluation Benchmarks for Chinese Accounting Reasoning}

Evaluation benchmarks for Chinese accounting reasoning mainly originate from two sources. The first includes general Chinese natural language inference benchmarks translated from English. The second consists of datasets derived from Chinese accounting examination standards or professional qualification exams. Compared with multi-step mathematical reasoning evaluation, Chinese accounting reasoning evaluation faces greater challenges due to differences in accounting standards, language structure, and task formulation.

The Chinese Language Understanding Evaluation (CLUE) benchmark \citep{xu2020} is a widely used Chinese counterpart to GLUE \citep{wang2018}. CLUE includes multiple natural language understanding tasks such as text classification, reading comprehension, and natural language inference, enabling comprehensive evaluation of Chinese language understanding. However, CLUE primarily targets general language understanding and does not explicitly address accounting reasoning.

Within CLUE, the Chinese Natural Language Inference (CMNLI) dataset evaluates whether a hypothesis is entailed, contradicted, or neutral given a premise. OCNLI is a native Chinese natural language inference dataset that better reflects Chinese linguistic and reasoning patterns. ACMC adapts CMNLI and OCNLI to financial and accounting contexts, but still lacks explicit accounting computation tasks.

To address these limitations, some studies explore evaluation using professional accounting examination questions, such as the Chinese Certified Public Accountant (CPA) exams. These exams cover accounting treatment selection, financial calculation, auditing judgment, and tax computation. However, CPA questions often involve lengthy contexts and complex domain knowledge, making them challenging for LLMs. Moreover, evaluation results are highly sensitive to prompt design, and model failures may result from misunderstanding rather than insufficient reasoning capability.

In summary, existing benchmarks either focus on general reasoning without accounting specificity or include accounting content without systematic reasoning evaluation. There remains a clear need for accounting-oriented reasoning benchmarks that integrate structured numerical computation with domain knowledge. This work addresses this gap by constructing accounting-specific evaluation datasets and systematically analyzing LLM performance across multiple reasoning dimensions.

\section{Experiments}

\subsection{Experimental Setup}

This study evaluates the accounting reasoning capabilities of large language models (LLMs). We first conceptualize accounting reasoning as the integration of logical reasoning and mathematical reasoning under domain-specific constraints. Based on this definition, we design evaluation tasks that cover multiple accounting scenarios and reasoning patterns, and construct corresponding benchmark datasets. The evaluation framework focuses on reasoning accuracy, reasoning consistency, and error propagation behavior, and adopts standardized evaluation procedures to ensure fair comparison across models.

Following common practices in prior work, we treat accounting reasoning as a composite capability that requires models to perform numerical computation, apply accounting rules, and maintain logical consistency across multi-step inference. Therefore, this study evaluates LLM performance from multiple perspectives, including multi-step calculation ability, accounting knowledge understanding, and comprehensive reasoning performance in realistic scenarios.

Empirical studies from the Stanford Alpaca team indicate that using curated datasets at approximately 5 percent of the original size can significantly improve model performance \citep{xia2024}. High-quality data plays a critical role in enhancing both reasoning accuracy and robustness. Accordingly, the first step of our evaluation is to assess the baseline reasoning capabilities of LLMs using general-purpose reasoning benchmarks, followed by specialized evaluation on accounting-oriented datasets.

To systematically measure accounting reasoning performance, we divide the evaluation into three categories: mathematical reasoning ability, accounting knowledge reasoning ability, and comprehensive accounting reasoning ability. These categories correspond to increasing levels of difficulty and realism in accounting tasks, ranging from basic numerical computation to complex scenario-based analysis.

\begin{table}[t]
\centering
\small
\caption{Benchmark Datasets and Reasoning Tasks}
\label{tab:benchmark_tasks}
\begin{tabular}{p{0.3\linewidth} p{0.58\linewidth}}
\toprule
\textbf{Reasoning Task} & \textbf{Datasets} \\
\midrule
Mathematical and Computational Reasoning 
& GSM8K, SVAMP, ASDiv, AQuA-RAT, MAWPS, AddSub, MultiArith, SingleEq, SingleOp \\
\midrule
Logical Reasoning 
& ProofWriter, EntailmentBank, RuleTaker, CLUTRR, FLD \\
\midrule
Commonsense Reasoning 
& CommonsenseQA, StrategyQA, ARC, SayCan, BoolQ, HotpotQA, OpenBookQA, PIQA, WikiWhy, COPA \\
\midrule
Abstract Reasoning 
& Last Letter Concatenation, Coin Flip \\
\midrule
Other 
& SNLI, MultiNLI, HellaSwag, SQuAD, BIG-bench, SCAN, BBH \\
\bottomrule
\end{tabular}
\end{table}

\subsection{Evaluation of Mathematical Reasoning Ability}

Existing benchmarks are widely used to evaluate the mathematical reasoning abilities of LLMs, including natural language inference and multi-step arithmetic reasoning datasets \citep{srivastava2022,yu2023,chang2023}. However, most general-purpose benchmarks are not designed to capture the complexity of accounting computation, particularly under real-world constraints such as conditional logic and cumulative error propagation.

Among existing datasets, GSM8K \citep{cobbe2021} is one of the most widely used benchmarks for evaluating multi-step mathematical reasoning. It consists of 8.5K high-quality elementary math word problems that require two to eight reasoning steps, primarily involving arithmetic operations such as addition, subtraction, multiplication, and division. These problems resemble simplified accounting calculations, where each intermediate result must be correctly derived to obtain the final answer.

However, GSM8K contains a large number of short problems with limited reasoning depth. To better evaluate extended reasoning chains, MR-GSM8K \citep{zeng2024} is constructed by filtering GSM8K problems with short reasoning trajectories and retaining those that require longer multi-step computation. Problems with answers shorter than 300 words are removed, resulting in a dataset that emphasizes sustained reasoning and error accumulation.

After further filtering, 586 MR-GSM8K problems requiring at least three computation steps remain. This subset forms the basis of our multi-step calculation evaluation benchmark, referred to as the \textit{Multi-Calculation Benchmark}. These problems are particularly suitable for assessing the stability and reliability of LLMs in accounting-style numerical reasoning.

\subsection{Evaluation of Chinese Accounting Reasoning Ability}

The evaluation of Chinese accounting reasoning ability presents unique challenges due to differences in language structure, accounting standards, and professional terminology. Existing Chinese reasoning benchmarks mainly originate from two sources: translated general-purpose reasoning datasets and datasets derived from professional accounting examinations.

The Chinese Language Understanding Evaluation (CLUE) benchmark \citep{xu2020} is widely used to evaluate general Chinese language understanding and reasoning capabilities. It includes tasks such as text classification, reading comprehension, and natural language inference, and serves as the Chinese counterpart to GLUE \citep{wang2018}. However, CLUE primarily evaluates general reasoning rather than accounting-specific inference.

Within CLUE, the Chinese Natural Language Inference (CMNLI) dataset evaluates whether a hypothesis is entailed, contradicted, or neutral with respect to a given premise. OCNLI is a native Chinese natural language inference dataset that avoids translation bias and better reflects Chinese reasoning patterns. Although these datasets provide useful signals for logical reasoning, they lack explicit accounting computation and domain constraints.

To address this limitation, we further consider accounting-related datasets derived from professional qualification examinations, such as the Chinese Certified Public Accountant (CPA) exams. CPA exam questions cover accounting treatment selection, financial calculation, auditing judgment, and tax computation. These tasks require models to integrate numerical reasoning with domain knowledge and regulatory constraints.

However, CPA-based evaluation also introduces challenges. Exam questions often contain long contextual descriptions, complex terminology, and implicit assumptions, which may lead to model failures due to misunderstanding rather than insufficient reasoning capability. Moreover, evaluation results can be sensitive to prompt design, making it difficult to isolate reasoning performance from language comprehension.

\subsection{Comprehensive Accounting Reasoning Evaluation}

To achieve a more balanced and reliable evaluation, this study combines general reasoning benchmarks, accounting-specific calculation tasks, and professional exam-style questions. By analyzing model performance across these dimensions, we aim to identify strengths and weaknesses in LLM accounting reasoning capabilities.

Our evaluation reveals that while LLMs demonstrate strong performance in basic numerical computation and short reasoning chains, their accuracy degrades significantly as reasoning depth increases. Error propagation across intermediate steps remains a major challenge. Furthermore, models often struggle to consistently apply accounting rules in complex scenarios, particularly when multiple constraints interact.

Overall, the experimental results indicate that current LLMs possess promising foundational reasoning abilities but still fall short of the requirements for real-world accounting applications. Improving accounting reasoning performance will require not only better training data and prompt strategies, but also deeper integration of domain knowledge and reasoning-aware model design.

\begin{figure}[t]
  \centering
  \includegraphics[width=0.9\linewidth]{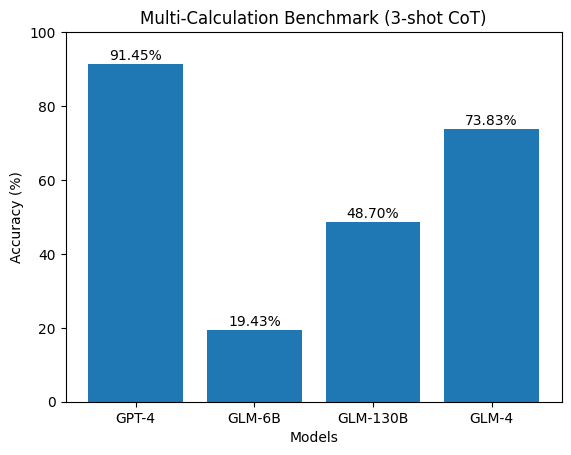}
  \caption{Multi-Calculation Benchmark: 3-shot Chain-of-Thought Evaluation Results.}
  \label{fig:2}
\end{figure}

\subsection{Evaluation Results and Analysis}

The experiments evaluate the performance of OpenAI GPT-4 and compare it with GLM-6B, GLM-130B, and GLM-4. During evaluation, Chain-of-Thought (CoT) prompting is used to encourage models to explicitly output intermediate reasoning steps, thereby improving reasoning performance and enhancing result interpretability. At the same time, Few-shot Learning is adopted by providing several example analyses to guide the models in learning how to reason and generate answers.

First, we conduct experiments under different prompt engineering settings, including Zero-shot, Few-shot, CoT, and Zero-shot-CoT. Experimental results show that outputs under Zero-shot and Zero-shot-CoT often contain logical inconsistencies. For example, models may produce a final answer but fail to ensure consistency between intermediate reasoning steps and the conclusion. As a result, relying solely on direct answer comparison when evaluating LLM outputs may lead to increased error rates. To mitigate this issue, we adopt Few-shot-CoT prompting in evaluation. Compared with Zero-shot approaches, Few-shot prompting significantly improves answer accuracy, particularly for multi-step reasoning tasks. Experimental statistics show that under Few-shot-CoT, model accuracy improves by approximately 50 percent. In accounting reasoning tasks, this approach enables models to correctly answer questions involving around nine arithmetic operations, demonstrating enhanced reasoning capability.

\begin{figure}[t]
  \centering
  \includegraphics[width=0.9\linewidth]{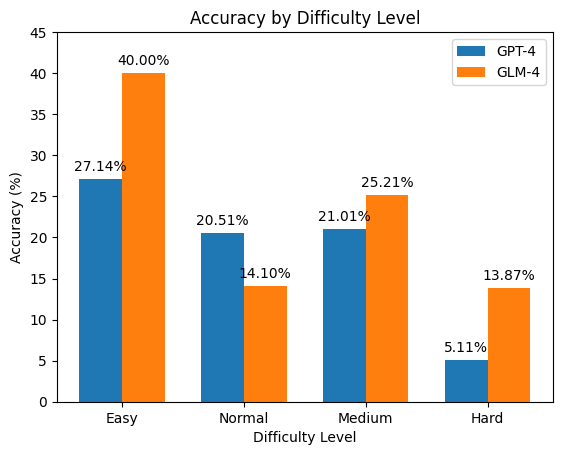}
  \caption{Accounting-Reasoning-Benchmark (3-shot CoT) Evaluation Results.}
  \label{fig:3}
\end{figure}

\begin{figure*}[t]
  \centering
  \includegraphics[width=0.9\linewidth]{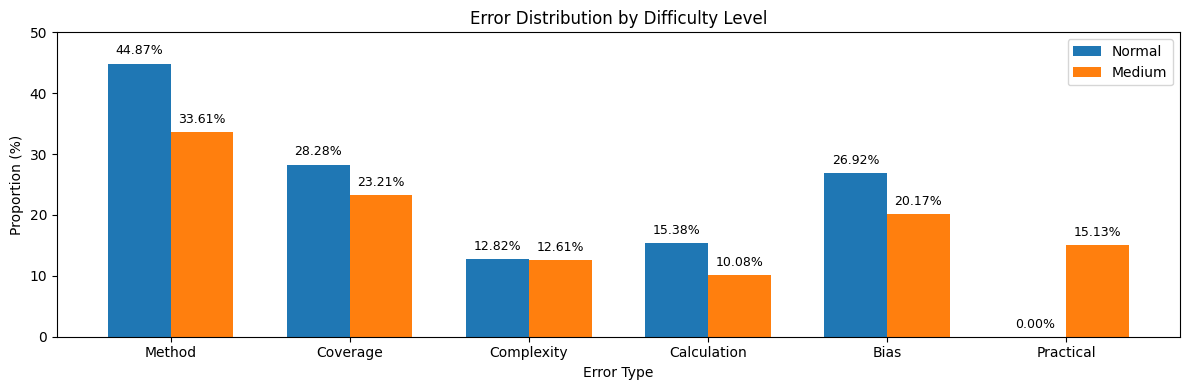}
  \caption{Comparison of Error Type Proportions between Normal and Medium Difficulty Levels.}
  \label{fig:4}
\end{figure*}

Based on these findings, we adopt Few-shot-CoT as the standard evaluation paradigm. Specifically, we use the GSM8K-style Few-shot-CoT template to structure model outputs. The reasoning process and final answer are separated explicitly. For example, given an accounting question involving asset purchases and depreciation calculations, the model first performs step-by-step reasoning and then outputs the final numerical result. This structure improves both reasoning clarity and evaluation consistency.

After obtaining model responses on the dataset, we perform quantitative evaluation by comparing the model outputs with ground truth answers. GLM-4 is used as the automatic evaluator to judge the correctness of responses from GLM-6B, GLM-130B, and GPT-4. Final accuracy is computed across all questions, and the results are summarized in the Multi-Calculation Benchmark evaluation.

GPT-4 achieves the highest overall accuracy in accounting reasoning tasks as of February 2024, establishing a strong performance benchmark among evaluated LLMs. However, GPT-4 still fails on certain complex accounting problems, indicating that further optimization is required. Evaluation results show that multi-step arithmetic reasoning remains challenging for smaller models such as GLM-6B, where accuracy remains around 20 percent. GLM-130B achieves approximately 60 percent accuracy on multi-step arithmetic tasks, which is still insufficient for professional accounting applications that typically require accuracy above 90 percent. Notably, although GLM-4 demonstrates strong performance, its accuracy remains below the expected threshold, highlighting the limitations of current LLMs in precise numerical reasoning.

Given that the benchmark focuses on multi-step arithmetic accounting problems, only GPT-4 and GLM-4 are further evaluated on the Accounting Reasoning Benchmark. Experimental results indicate that under accounting-specific reasoning tasks, GPT-4 achieves an accuracy of 16.58 percent, while GLM-4 achieves 21.78 percent. This suggests that GLM-4 slightly outperforms GPT-4 in accounting knowledge reasoning. However, as task difficulty increases with the number of calculation steps, the performance gap widens. Neither model demonstrates fully reliable accounting reasoning ability.

Importantly, both models fail to reach human-level performance in accounting knowledge and reasoning. GLM-4 shows advantages in structured accounting reasoning, while GPT-4 demonstrates better robustness on general reasoning tasks. As the number of calculation steps increases, both models exhibit declining accuracy. For the most complex accounting questions, accuracy drops to around 40 percent. There is no direct linear relationship between reasoning depth and accuracy, indicating that GLM-4 struggles to adapt to increasingly complex accounting scenarios. These findings suggest that although LLMs can assist in accounting tasks, they cannot yet replace human expertise.

Further error analysis categorizes model failures into several types. First, accounting logic misuse occurs when models apply incorrect accounting principles during reasoning. Examples include misclassification of asset treatment, misunderstanding of accounting standards, and incorrect application of tax regulations. These errors indicate insufficient understanding of accounting logic.

Second, incomplete knowledge coverage arises when models fail to fully capture domain-specific accounting knowledge. In some cases, models overlook critical accounting conditions or constraints, resulting in incorrect conclusions.

Third, complex multi-branch reasoning errors occur in problems requiring multiple intertwined accounting procedures. LLMs struggle to track dependencies across multiple steps, especially when different accounting treatments interact.

Fourth, arithmetic and logical inconsistencies are observed when models produce correct reasoning structures but make numerical calculation errors. These errors reflect weaknesses in precise arithmetic execution.

Fifth, conceptual ambiguity appears when models demonstrate vague understanding of accounting concepts. Although reasoning steps appear coherent, underlying assumptions are flawed due to unclear conceptual grounding.

Sixth, principle-level misunderstanding occurs when models misinterpret foundational accounting principles, even when factual details are correct. These errors reveal limitations in deep conceptual understanding rather than surface-level mistakes.

A more detailed classification shows that the most frequent errors involve misunderstanding of accounting principles and insufficient conceptual coverage. Together, these account for over 50 percent of total errors. Errors related to detailed operational procedures, such as bookkeeping entries and financial statement preparation, are less frequent but still significant. This distinction highlights that while LLMs perform reasonably well on procedural tasks, they struggle with principle-driven accounting reasoning.

Further comparison between error types shows that principle-level errors are more severe than procedural errors, as they often lead to cascading mistakes throughout the reasoning process. In contrast, procedural errors tend to affect only local steps. This pattern aligns with the structure of professional accounting exams, where conceptual understanding is critical for advanced problem solving.

Overall, the results demonstrate that current LLMs lack robust accounting reasoning capabilities. While they show promise in assisting with accounting-related tasks, significant improvements are required in domain knowledge integration, reasoning stability, and numerical accuracy before they can be reliably deployed in professional accounting environments.

\section{Conclusion}

This work investigates the reasoning capabilities of large language models (LLMs) in the accounting domain using two evaluation benchmarks. Experimental results show that while LLMs perform well on general multi-step reasoning tasks, their accuracy drops significantly on accounting-specific reasoning, indicating a clear gap between general reasoning ability and domain expertise. Error analysis further reveals that current LLMs struggle to apply accounting principles consistently in complex reasoning scenarios. These findings suggest that LLMs are not yet ready for reliable deployment in professional accounting applications.

\section*{Limitations}

This study has several limitations. First, the benchmark focuses on structured accounting reasoning tasks and does not fully capture the diversity and complexity of real-world accounting practices. Second, only a limited number of representative LLMs are evaluated, and the results may not generalize to models trained with different architectures or domain-specific data. Third, the evaluation is based on prompt-based reasoning paradigms, and model performance may be sensitive to prompt design choices. Addressing these limitations will require broader benchmarks, more diverse models, and systematic domain adaptation in future work.

% Bibliography entries for the entire Anthology, followed by custom entries
%\bibliography{anthology,custom}
% Custom bibliography entries only
\bibliography{custom}

\end{document}